\documentclass[lettersize,journal]{IEEEtran}
\usepackage[T1]{fontenc}
\usepackage[utf8]{inputenc}
\usepackage{amsmath,amssymb}
\usepackage{graphicx}
\usepackage{cite}
\usepackage{xcolor}
\usepackage{url}
\usepackage[hidelinks]{hyperref}
\newcommand{\meanabs}[1]{\left\langle\left|#1\right|\right\rangle}
\newcommand{\method}{AstraSR}
\begin{document}
\bstctlcite{IEEEexample:BSTcontrol}
\title{AstraSR: Real-World Thermal Super-Resolution with GPT-6 Astra}
\author{Mengyuan Li$^{1}$, Changhong Fu$^{1,*}$, Jun Zhang$^{2}$, Ziyu Lu$^{1}$, Yuhang Zhang$^{1}$, and Haobo Zuo$^{3}$
\thanks{$^{1}$Mengyuan Li, Changhong Fu, Ziyu Lu, and Yuhang Zhang are with the School of Mechanical Engineering, Tongji University, Shanghai 201804, China.
{\tt\small changhongfu@tongji.edu.cn}}
\thanks{$^{2}$Jun Zhang is with the School of Electrical and Electronic Engineering, Nanyang Technological University, Singapore.}
\thanks{$^{3}$Haobo Zuo is with the School of Computing and Data Science, the University of Hong Kong, Hong Kong 999077, China.}
\thanks{\hspace{0.3cm}*Corresponding author}
}
\maketitle
\begin{abstract}
  Real-world thermal super-resolution (SR) is constrained by limited sensor resolution and the difficulty of obtaining corresponding high-resolution (HR) observations for direct model supervision. Conventional SR methods typically construct training pairs by treating captured thermal images with real-world degradations as HR references and applying predefined degradation to generate synthetic low-resolution (LR) inputs. Such a construction not only introduces a domain gap between synthetic and captured LR observations but also retains acquisition degradations in the supervision. To address this issue, we propose AstraSR, a real-world thermal SR method guided by GPT-6 Astra, a frontier multimodal generative model endowed with emergent and transformative visual capabilities. Specifically, we construct a dataset of image pairs by using captured LR thermal images to condition GPT-based HR reference. We develop a direct generative supervision strategy that learns from captured thermal inputs paired with GPT-generated HR references. Pixel, gradient, and perceptual losses jointly supervise the transfer of intensity patterns, structural boundaries, and visual details from the generated references. Qualitative comparisons with seven existing state-of-the-art real-world SR methods show continuous object contours, distinct structural boundaries, and smooth intensity transitions in the thermal scenes. These results demonstrate that AstraSR outperforms existing real-world SR methods in both thermal clarity and structural coherence.
    
    \end{abstract}
\section{Introduction}
\label{sec:introduction}
\IEEEPARstart{T}{hermal} imaging provides useful scene information when visible illumination is limited or thermal contrast is informative. Applications such as nighttime observation, search and rescue, and infrastructure inspection benefit from being able to distinguish references and their surrounding structures~\cite{edgesr,anytsr++}. However, the available spatial detail is constrained by the thermal sensor and imaging optics. During image acquisition, motion, vibration, and sensor noise can further obscure boundaries that already occupy few pixels. Enlarging such images by interpolation increases their display size but does not resolve the underlying ambiguity. Therefore, learning to recover reliable high-resolution (HR) details from real captured thermal observations is essential for practical thermal super-resolution (SR).

\begin{figure}[!t]
\centering
\setlength{\abovecaptionskip}{0.cm}
\includegraphics[width=\linewidth]{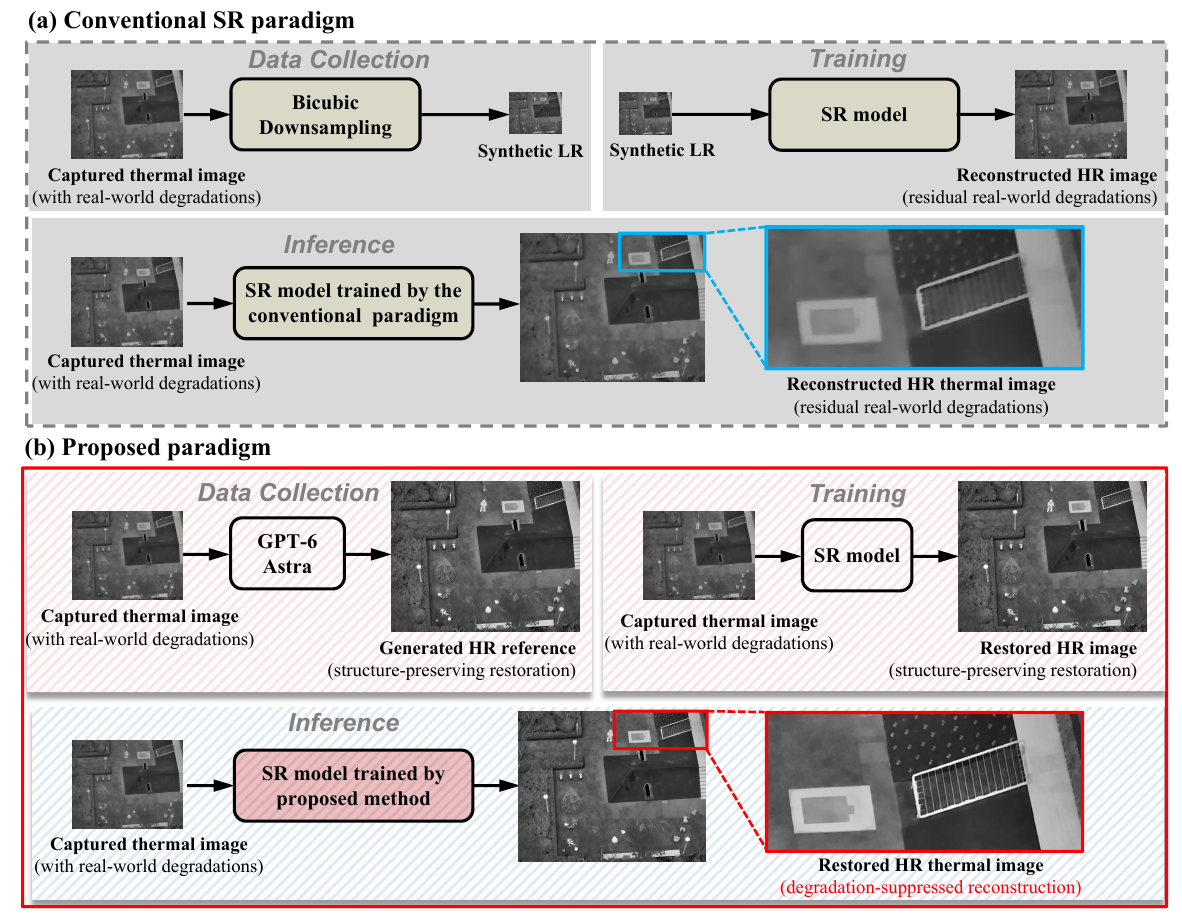}
\caption{Comparison of two thermal super-resolution paradigms. (a) The conventional paradigm constructs synthetic LR inputs by degrading captured thermal images that are used as HR references, causing a mismatch with real captured LR observations while retaining acquisition degradations in the supervision. (b) AstraSR leverages the emergent and transformative visual capabilities of GPT-6 Astra to construct structure-preserving HR references directly from captured thermal inputs, which are then used to supervise SR training for degradation-suppressed reconstruction. Enlarged regions show representative differences in boundary and structural recovery.}
\label{fig:paradigm}
\vspace{-10pt}
\end{figure}

Supervised single-image SR learns a mapping from a low-resolution (LR) image to a HR reference. Convolutional models established the effectiveness of learning this mapping directly, and subsequent residual and attention-based networks increased its representational capacity~\cite{7115171,edsr,rcan}. In a common training formulation, a predefined operator produces the LR input from an HR image, with bicubic downsampling providing a convenient and reproducible example. As illustrated in Fig.~\ref{fig:paradigm}(a), applying this construction to captured thermal images makes the original image the HR reference and its downsampled version the LR input. Noise and blur in that original image are therefore also present in the reference. Although this formulation yields well-aligned training pairs, the learned mapping remains constrained by the assumed degradation process and the quality of the captured reference. Hence, it is imperative to construct training pairs that faithfully preserve the characteristics of captured thermal inputs throughout training~\cite{Perceptual,wang2018esrgan}.

Real-world SR extends this formulation to more complex and unknown input degradations. Existing approaches generally improve reconstruction robustness by broadening synthetic degradation distributions~\cite{bsrgan,real_esrgan}, incorporating learned high-quality image priors~\cite{swinir,femasr}, or exploiting generative models for perceptual reconstruction~\cite{resshift,sinsr,pisasr,chen2025adcsr}. These advances substantially broaden the available approaches to processing real photographs. Despite this progress, transferring such methods to real-world thermal imagery presents unresolved difficulties. First, even a broad synthetic degradation distribution may not reproduce the joint image characteristics produced by a particular thermal camera under real acquisition conditions. Second, the released models considered here were developed primarily with natural-image training data, whose appearance differs from the grayscale intensity patterns of thermal imagery. Larger generative priors do not by themselves establish that an output detail is supported by a thermal observation. Third, generating LR inputs from an HR source changes the training input distribution: the network learns from simulated observations rather than the measurements it will receive in use. Direct supervision with captured LR images would avoid this particular substitution, but obtaining aligned sensor-captured HR thermal references remains difficult. This setting calls for a teacher that can interpret the scene structure contained in a degraded thermal observation while generating a higher-resolution representation under explicit restoration instructions. GPT-6 Astra~\cite{gpt6astra}, as a frontier general-purpose multimodal model, combines visual understanding, instruction following, and image generation within a unified framework, making it well suited to this role. Rather than synthesizing an independent scene from text alone, Astra can condition generation on the captured thermal observation and use its broad visual priors to recover structures that are weakly represented at the sensor resolution~\cite{instructpix2pix}. This capability motivates our use of Astra as an offline generative teacher for constructing HR supervision from captured thermal inputs.

To address these challenges, we propose \method, a GPT-guided method for real-world thermal SR. The main contributions are as follows:
\begin{itemize}
  \item We propose \method, which transfers the visual detail provided by a generative teacher to a thermal SR model. By using GPT only during offline reference construction, \method\ enables direct inference on captured thermal images without querying the generative model.
  \item We construct a real-world thermal SR dataset using GPT-6 Astra, pairing captured LR thermal images with corresponding GPT-generated HR images. These pairs provide supervision for learning fine structures and object boundaries from real thermal observations.
  \item We develop a direct real-to-teacher training strategy that retains captured images as LR inputs without synthetic downsampling. The strategy combines pixel, gradient, and perceptual losses to learn the teacher-provided intensity patterns, local contours, and visual details for thermal SR.
\end{itemize}
\section{Related Work}
\label{sec:related}

\subsection{Real-World Image Super-Resolution}
Real-world SR addresses inputs for which the degradation is not known in advance. A prominent approach broadens the synthetic training distribution. BSRGAN~\cite{bsrgan} shuffles blur, resizing, and noise operations, whereas Real-ESRGAN~\cite{real_esrgan} models repeated degradation and includes ringing and overshoot effects. However, expanding a synthetic degradation pipeline does not ensure that its training inputs reproduce the sensor-dependent characteristics of real captured thermal images.
Learned high-quality image priors complement degradation modeling. FeMaSR~\cite{femasr} represents such a prior with a vector-quantized codebook and reconstructs images by matching degraded features to codebook entries. Real-world variants of SwinIR~\cite{swinir} use an attention-based backbone within a perception-oriented training setting. Diffusion methods provide a further family of real-world SR models. ResShift~\cite{resshift} constructs a short diffusion process around the residual between LR and HR representations. SinSR~\cite{sinsr} learns a deterministic single-step mapping from a diffusion teacher. PiSA-SR~\cite{pisasr} separates pixel-level and semantic-level adaptation into two LoRA components, while AdcSR~\cite{chen2025adcsr} uses adversarial distillation to compress a pretrained diffusion model. Nevertheless, stronger natural-image priors alone do not ensure that reconstructed textures and boundaries remain consistent with the content of a thermal observation. 

\subsection{Large Generative Models for Image Generation}
Large generative models have expanded the ways in which language and images can condition image synthesis. Latent diffusion models~\cite{rombach2022ldm} perform generation in a learned compressed representation, supporting HR synthesis with flexible conditioning. Work on DALL-E~3~\cite{openai2023dalle3} shows that improved training captions can strengthen correspondence between instructions and generated content. GPT-4o image generation~\cite{gpt4oimage} further illustrates an interface that combines language instructions, supplied images, and iterative image editing. 
Compared with task-specific generative models, GPT-6 Astra~\cite{gpt6astra} provides a more general multimodal capability that integrates visual understanding, instruction following, and image-conditioned generation. This combination makes it particularly suitable as a generative teacher for thermal SR, where weak scene structures must first be interpreted before a detailed HR reference can be produced.
Accordingly, GPT-6 Astra offers a suitable means of constructing task-specific HR supervision for thermal SR.

\section{Proposed Method}
\label{sec:method}
\begin{figure*}
\centering
\includegraphics[width=0.95\textwidth]{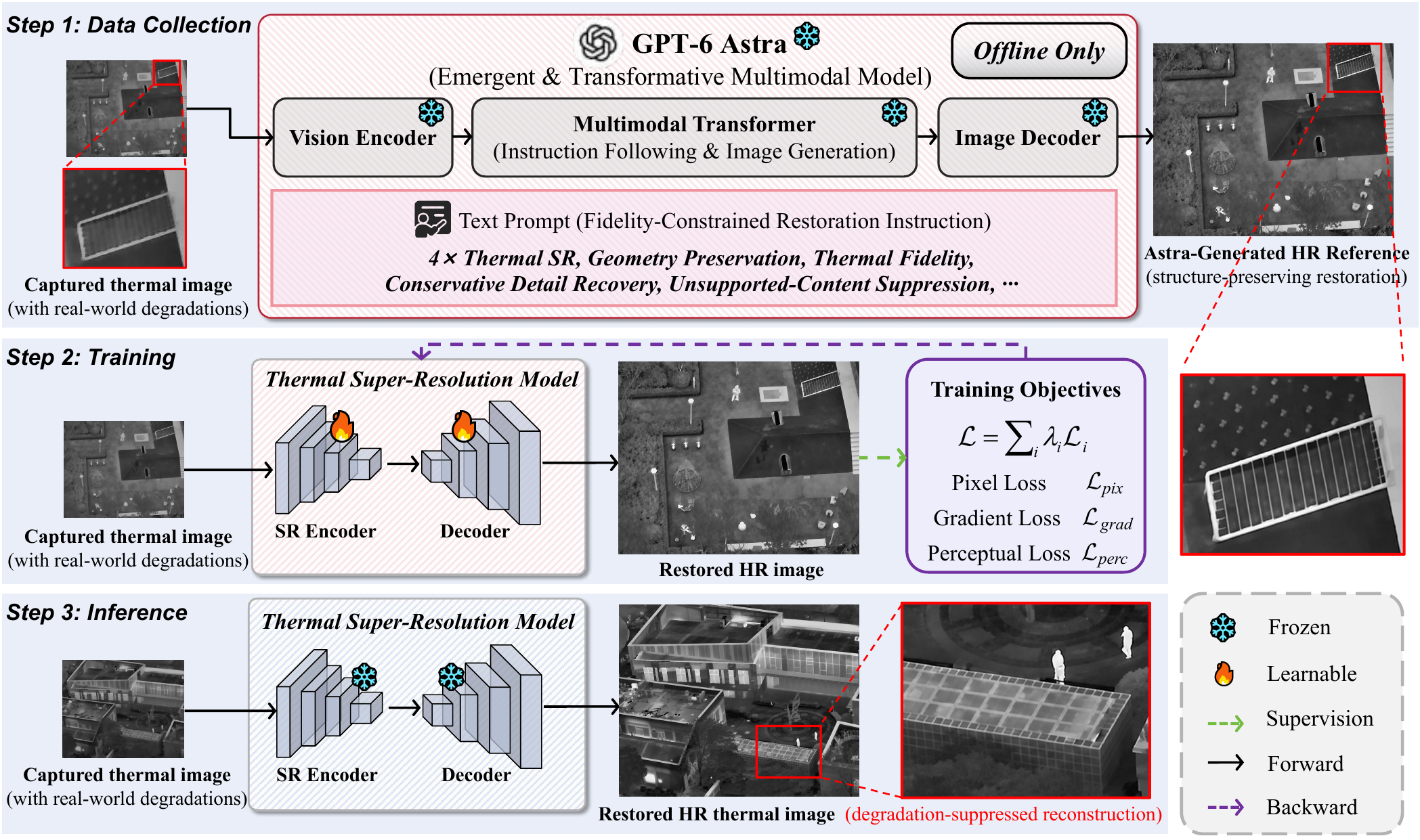}
\caption{Overview of the proposed AstraSR framework. GPT-6 Astra is employed only offline to construct structure-preserving HR references from captured thermal inputs under fidelity-constrained restoration instructions. These references supervise the learnable thermal SR model using pixel, gradient, and perceptual objectives. After training, the SR model is frozen and independently performs degradation-suppressed thermal SR, without requiring GPT-6 Astra during inference.}
\label{fig:method}
\vspace{-10pt}
\end{figure*}
\subsection{Problem Definition and Method Overview}
\label{sec:overview}
Let $x\in[0,1]^{1\times H\times W}$ denote a real captured thermal image, represented as one grayscale channel. We consider a fixed scale $s$ and seek an output $\hat y\in\mathbb{R}^{1\times sH\times sW}$ from a parameterized mapping
\begin{equation}
  \hat y=G_\theta(x)\quad.
  \label{eq:prediction}
\end{equation}

The supervision is a generated reference $y^T$, where superscript $T$ denotes the teacher. We write:
\begin{equation}
  y_i^T=\mathcal{T}(x_i,q_i)\quad,
  \label{eq:teacher}
\end{equation}
where $\mathcal{T}$ denotes the offline GPT-based reference-construction procedure and $q_i$ collects its instructions and generation settings. Conventional synthetic-pair training constructs an input $\tilde x_i=\mathcal{D}(y_i)$ from an HR image through a degradation operator $\mathcal{D}$. Our training set instead consists of $\mathcal{S}=\{(x_i,y_i^T)\}_{i=1}^{N}$, where each $x_i$ is the captured image itself. The estimation problem is:
\begin{equation}
  \theta^\star=\arg\min_\theta\frac{1}{N}
  \sum_{i=1}^{N}\mathcal{L}\bigl(G_\theta(x_i),y_i^T\bigr)\quad.
  \label{eq:objective}
\end{equation}

The overall framework of AstraSR is presented in Fig.~\ref{fig:method}, illustrating how the proposed learning formulation is realized through offline HR reference construction, supervised SR training, and independent inference. In step 1, the captured thermal image is processed by GPT-6 Astra under a fidelity-constrained restoration instruction. Leveraging its multimodal generative capabilities, GPT-6 Astra constructs a structure-preserving HR reference while maintaining scene geometry and thermal fidelity, encouraging conservative detail recovery and suppressing unsupported content generation. This stage is performed offline only, and the generated HR reference is subsequently used as supervision rather than participating in inference. In step 2, a learnable thermal SR model takes the captured thermal image as input and predicts a dense HR reconstruction. The prediction is optimized against the generated HR reference using pixel, gradient, and perceptual losses, with gradients propagated only through the SR model. After training, the SR model is frozen and directly restores captured thermal inputs into degradation-suppressed HR thermal images in step 3.

\noindent\textbf{Remark 1:} The central design of \method\ connects offline GPT-based reference generation with a separately trained thermal SR model. This makes generative supervision available for learning from real captured inputs while allowing the trained model to perform SR without online GPT queries.

\subsection{GPT-Guided HR Thermal Image Generation}
\label{sec:construction}
We use GPT-6 Astra to construct HR training references from existing real captured thermal images. The source image provides the scene content, while text instructions guide the generation toward clearer boundaries and more visible fine structures. The aim is to produce an HR representation of the same thermal scene that can serve as supervision for the SR model. Generation focuses on structures that are difficult to distinguish in the captured input, including thin lines, small foreground objects, and repeated geometric patterns. The desired references combine more distinct local contours with smoother intensity transitions in broad regions, while retaining the overall organization of the source scene. Each generated HR reference is further examined image by image against its corresponding captured LR input to verify that the recovered details remain consistent with the observable scene structure and do not introduce unsupported content. These properties motivate the pixel, gradient, and perceptual objectives introduced in Section~\ref{sec:training}. All references are generated offline and paired with their original LR images, so the resulting dataset can be reused during training without further GPT queries.

\begin{figure*}
\centering
\includegraphics[width=1\textwidth]{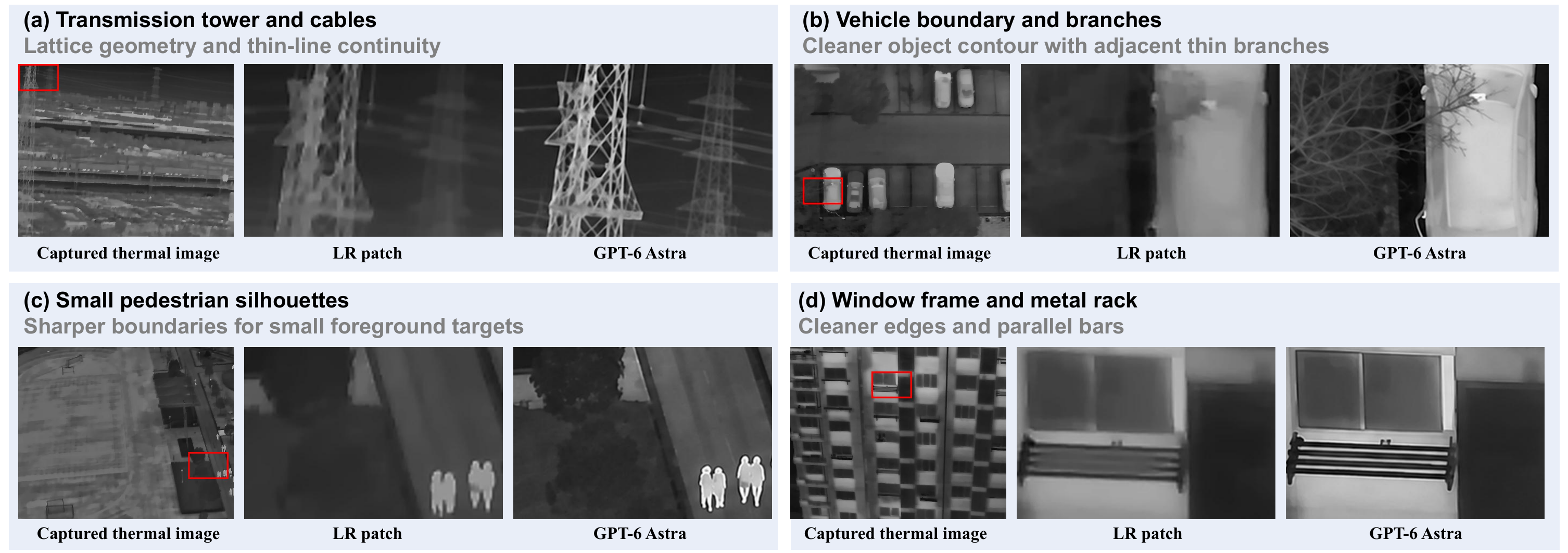}
\caption{Examples of captured thermal inputs and GPT-generated HR references. Panels (a) and (b) show a transmission tower with cables and vehicle windows with roof contours, respectively. Panels (c) and (d) show small pedestrian silhouettes and a window frame with a metal rack, respectively. These examples illustrate the contour and structural detail represented in the generated training references.}
\label{fig:dataset}
\vspace{-10pt}
\end{figure*}

Fig.~\ref{fig:dataset} illustrates the structural detail represented in the GPT-generated HR references. In (a), the tower lattice and cables appear more distinctly delineated than in the enlarged LR crop. In (b), the window and sunroof boundaries are more clearly defined, while the vehicle roof contours appear smoother and more continuous. In (c), neighboring pedestrian silhouettes have sharper outlines and more distinct boundaries between adjacent targets. In (d), the window frame and the horizontal bars beneath it exhibit cleaner edges and more visible gaps. These generated references provide training targets emphasizing thin structures, object contours, small targets, and repeated geometric patterns.

\noindent\textbf{Remark 2:} Existing captured images provide the LR observations, while image-conditioned GPT generation supplies detailed HR supervision without requiring corresponding measurements from a higher-resolution thermal sensor.

\subsection{Direct Real-to-Teacher Training}
\label{sec:training}
\subsubsection{Direct Pair Learning}
The input-reference pair presented to the model is $(x_i,y_i^T)$. No bicubic reduction of $y_i^T$, synthetic blur, added noise, or simulated compression is used to construct a substitute LR image. Spatial cropping and a shared horizontal flip provide training samples while retaining the content of the captured input.

% This choice exposes the model to the observed thermal input distribution, but it does not make the supervision error-free. A discrepancy between $x_i$ and $y_i^T$ is learned according to the training objective, whether it represents a desirable visual change or an unwanted teacher modification. We therefore define the objective explicitly as prediction of the HR reference. The present experiments examine this formulation as a whole. They do not isolate the causal effect of direct pairing from all other training choices.

\subsubsection{Pixel Loss}
For a mini-batch of $B$ pairs, let $\hat y_i=G_\theta(x_i)$. We use the mean absolute pixel distance:
\begin{equation}
  \mathcal{L}_{\mathrm{pix}}=
  \frac{1}{B}\sum_{i=1}^{B}\meanabs{\hat y_i-y_i^T}\quad,
  \label{eq:pixel}
\end{equation}
where $\langle\cdot\rangle$ denotes the arithmetic mean over all entries of its argument.

\subsubsection{Gradient Loss}
To complement the pixel objective, we compare horizontal and vertical image derivatives computed with fixed Sobel kernels,
\begin{equation}
  K_x=\frac{1}{8}
  \begin{bmatrix}-1&0&1\\-2&0&2\\-1&0&1\end{bmatrix}\quad,
  \qquad K_y=K_x^{\mathsf T}\quad.
  \label{eq:sobel}
\end{equation}

Writing $\nabla_d z=K_d*z$ for direction $d$, the loss is
\begin{equation}
  \mathcal{L}_{\mathrm{grad}}=
  \frac{1}{2B}\sum_{i=1}^{B}\sum_{d\in\{x,y\}}
  \meanabs{\nabla_d\hat y_i-\nabla_d y_i^T}\quad.
  \label{eq:gradient}
\end{equation}

Convolution uses replicated boundary padding. The term encourages agreement with the local intensity transitions and contours in the teacher-generated reference. It is a reference-gradient objective, rather than a constraint that forces the enlarged output to reproduce the noisy LR gradients.

\subsubsection{Perceptual Loss}
Perceptual supervision compares predictions and references in a frozen feature representation~\cite{Perceptual}. We use ImageNet-pretrained VGG-16, replicating each grayscale image across three channels before applying the standard ImageNet channel normalization. Let $\psi$ denote this preprocessing and $\phi_l$ the feature map at layer $l$. Then
\begin{equation}
  \begin{aligned}
  \mathcal{L}_{\mathrm{perc}}
    &=\frac{1}{B|\mathcal{F}|}\sum_{i=1}^{B}\sum_{l\in\mathcal{F}}d_l(\hat y_i,y_i^T)\quad,\\
  d_l(a,b)&=\meanabs{\phi_l(\psi(a))-\phi_l(\psi(b))}\quad,
  \end{aligned}
  \label{eq:perceptual}
\end{equation}
where $\mathcal{F}$ contains ReLU1\_2, ReLU2\_2, ReLU3\_3, and ReLU4\_3. Each layer contributes its mean absolute feature distance with equal layer weight. The feature network stays in evaluation mode and is not optimized. Gradients propagate through its prediction branch to the SR network. 

\begin{figure*}
\centering
\includegraphics[width=1\textwidth]{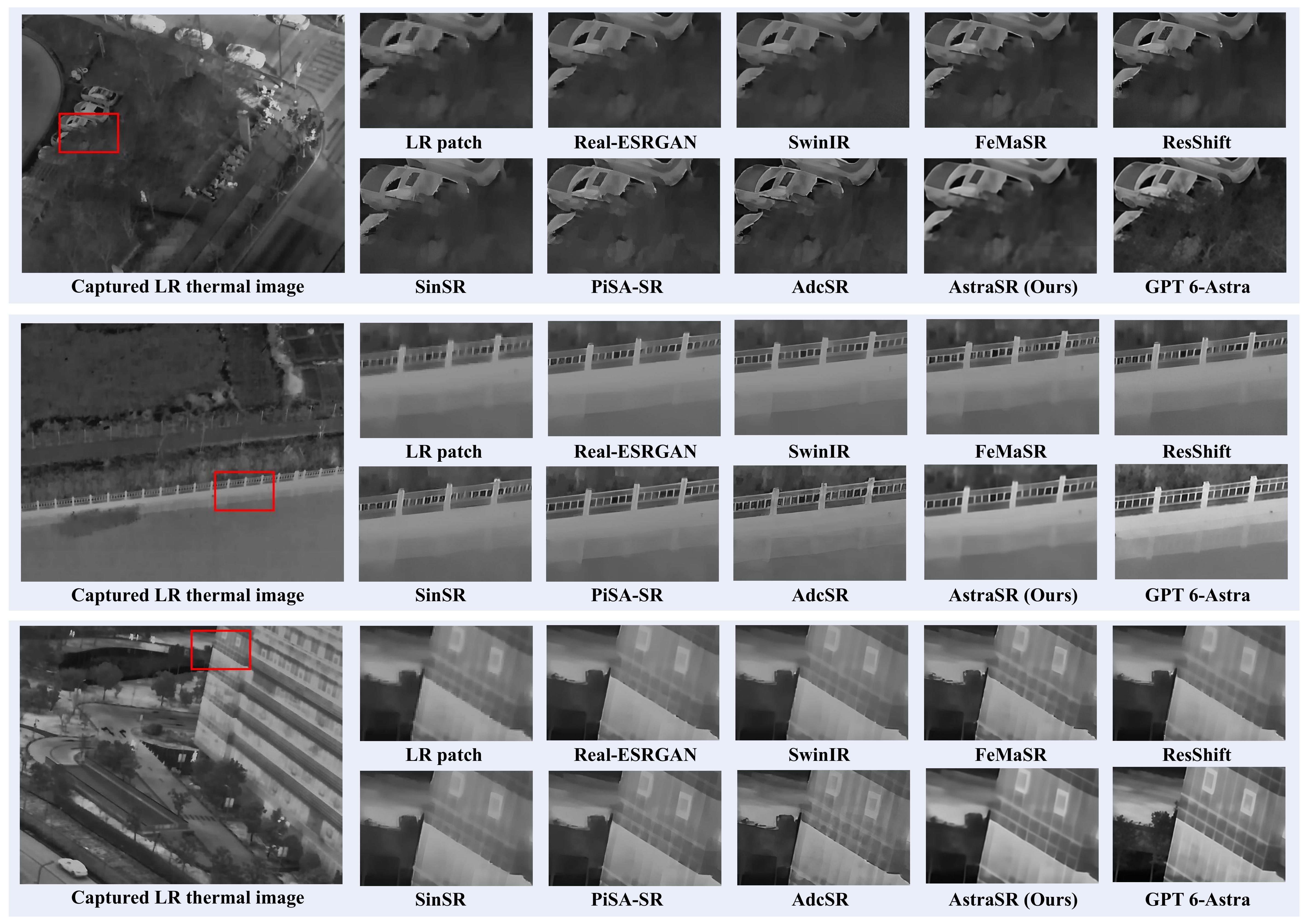}
\caption{Visual comparison of direct $4\times$ thermal super-resolution. The red box in each full-scene view identifies the region enlarged. The compared SOTA methods includes Real-ESRGAN~\cite{real_esrgan}, SwinIR~\cite{swinir}, FeMaSR~\cite{femasr}, ResShift~\cite{resshift}, SinSR~\cite{sinsr}, PiSA-SR~\cite{pisasr}, and AdcSR~\cite{chen2025adcsr}. The LR patch provides input context, whereas the others show the corresponding enlarged outputs. AstraSR achieves the strongest visual restoration among the compared methods, producing structurally coherent details and well-defined boundaries while closely reproducing the restoration characteristics of the GPT-6 Astra-generated HR reference.}
\label{fig:comparison}
\vspace{-10pt}
\end{figure*}

\subsubsection{Overall Objective}
Following the three supervision paths in Fig.~\ref{fig:method}, the complete objective is:
\begin{equation}
  \mathcal{L}_{\mathrm{total}}=
  \lambda_{\mathrm{pix}}\mathcal{L}_{\mathrm{pix}}+
  \lambda_{\mathrm{grad}}\mathcal{L}_{\mathrm{grad}}+
  \lambda_{\mathrm{per}}\mathcal{L}_{\mathrm{perc}}\quad,
  \label{eq:total}
\end{equation}
where weights $\lambda_{\mathrm{pix}}=1$, $\lambda_{\mathrm{grad}}=0.05$, and $\lambda_{\mathrm{perc}}=0.05$. Pixel supervision supplies the main reference constraint, while gradient and feature distances supplement it with local transitions and multilevel appearance information.

\noindent\textbf{Remark 3:} The training contribution is the direct use of real LR--generated HR pairs with complementary dense supervision. Applying pixel, gradient, and perceptual objectives to these pairs allows the model to learn reference intensity patterns, local boundaries, and visual detail while retaining the captured input distribution.

\section{Experiments}
\label{sec:experiments}
\subsection{Experimental Setup}
\label{sec:setup}
\subsubsection{Dataset Setup}
Experiments are conducted on the UAV-TSR dataset~\cite{anytsr}, which consists of 800 pairs for training and 300 captured LR images for testing. Each training pair contains a $640\times512$ thermal image and a $2560\times2048$ GPT-generated HR reference. Testing enlarges LR images directly to $2560\times2048$. No preliminary downsampling is applied. 

\begin{table}[!b]
\centering
\caption{Parameter and computational complexity. AstraSR achieves a favorable efficiency--complexity trade-off, requiring substantially fewer parameters and FLOPs than most recent real-world SR methods.}
\label{tab:complexity}
\small
\begin{tabular}{lrr}
\hline
Method & Params (M) & FLOPs (G) \\
\hline
Real-ESRGAN~\cite{real_esrgan} & 16.70 & 587.43 \\
SwinIR~\cite{swinir}& 28.01 & 1004.98 \\
FeMaSR~\cite{femasr} & 34.06 & 766.34 \\
ResShift~\cite{resshift} & 173.92 & 6509.61 \\
SinSR~\cite{sinsr} & 173.92 & 5293.94 \\
PiSA-SR~\cite{pisasr} & 949.57 & 4435.43 \\
AdcSR~\cite{chen2025adcsr} & 456.07 & 1099.00 \\
\hline
\textbf{AstraSR (Ours)} & \textbf{20.47} & \textbf{764.57} \\
\hline
\end{tabular}
\end{table}

\subsubsection{Implementation Details}
We adopt the encoder from AnyTSR~\cite{anytsr} and replace its arbitrary-scale upsampler with a fixed $4\times$ reconstruction module. The encoder contains 4 state-space groups, each with 8 blocks. The encoder feature width is set to 240 channels, and the reconstruction module uses 64 channels. 
Training uses random $128\times128$ LR patches from the real captured LR thermal images and corresponding $512\times512$ HR reference patches.
The batch size is 16. The model is trained for 100 epochs. Adam uses an initial learning rate of $10^{-4}$ and coefficients $(\beta_1,\beta_2)=(0.9,0.99)$. The learning rate is halved after epochs 20, 40, 60, and 80. An exponential moving average of generator parameters used decay 0.999. 

\subsection{Parameter and Computational Complexity}
\label{sec:complexity}
Table~\ref{tab:complexity} compares the parameter counts and estimated inference FLOPs of the evaluated models. All computations use an LR input of $128\times128$ pixels for $4\times$ SR. AstraSR contains 20.47M parameters and requires approximately 764.57 GFLOPs under this protocol. Its parameter count lies between those of Real-ESRGAN and SwinIR, and its estimated computational cost is close to that of FeMaSR. It has lower parameter and operation counts than the evaluated diffusion-based models, although Real-ESRGAN remains smaller and requires fewer operations.

\subsection{Qualitative Comparisons}
\label{sec:qualitative}
We compare \method\ with seven existing real-world super-resolution methods. These include Real-ESRGAN~\cite{real_esrgan}, SwinIR~\cite{swinir}, FeMaSR~\cite{femasr}, ResShift~\cite{resshift}, SinSR~\cite{sinsr}, PiSA-SR~\cite{pisasr}, and AdcSR~\cite{chen2025adcsr}. Fig.~\ref{fig:comparison} presents direct $4\times$ enlargement of captured thermal images, together with matched crops from the regions marked by red boxes. The examples cover object contours, repeated thin structures, and geometric boundaries, allowing comparison of both local detail and intensity transitions.  Real-ESRGAN and SwinIR improve edge definition, making major contours and straight lines easier to distinguish. FeMaSR produces more segmented intensity regions and additional local variation in some crops. ResShift and SinSR delineate the main structures, although several narrow openings and boundary transitions remain uneven or soft. PiSA-SR and AdcSR render more surface variation in parts of the displayed scenes, with some regions appearing more strongly textured or partitioned. These differences show that stronger local contrast does not always coincide with smoother surfaces or more regular fine structures. Across the displayed examples, \method\ combines continuous object contours and distinct structural boundaries with smoother intensity transitions across broad surfaces. Repeated elements and rectangular outlines remain visually prominent, while adjacent regions exhibit a more consistent appearance. This balance makes the main structures easier to distinguish without emphasizing texture throughout the image. These results demonstrate that AstraSR achieves more coherent structural reconstruction, with clearer boundaries, more continuous contours, and smoother intensity transitions than competing methods.

\section{Conclusion}
\label{sec:conclusion}
We presented \method\ for real-world thermal super-resolution and constructed a dataset pairing captured LR thermal images with GPT-6 Astra-generated HR references. By using captured images directly as inputs, the method avoids constructing LR training samples through synthetic downsampling. Our direct real-to-teacher strategy trains a single-channel SR model with pixel, gradient, and perceptual losses, enabling $4\times$ inference without querying GPT. Qualitative comparisons show clearer object contours and more distinct structural boundaries alongside smoother intensity transitions in the illustrated scenes. These observations support the use of GPT-generated HR supervision for learning the appearance of enhanced thermal images. Future work will explore an active learning paradigm where the lightweight SR model identifies challenging degradation cases and iteratively queries the offline Astra model with adaptive prompts, dynamically generating informative positive and negative training pairs to continuously evolve the training distribution.
\bibliographystyle{IEEEtran}
\bibliography{references}
\end{document}